\documentclass[runningheads]{llncs}

\usepackage{eccv}

\usepackage{eccvabbrv}

\usepackage{graphicx}
\usepackage{booktabs}

\usepackage[accsupp]{axessibility}  

\usepackage{textcomp}
\usepackage{stfloats}
\usepackage{url}
\usepackage{makecell}
\usepackage{verbatim}
\usepackage{graphicx}
\usepackage{cite}
\usepackage{balance}
\usepackage{booktabs}       
\usepackage{multirow}      
\usepackage{colortbl}      
\usepackage{xcolor}        
\usepackage{algorithm}
\usepackage{algorithmic}
\newcommand{\best}[1]{\textcolor{red}{\textbf{#1}}}
\newcommand{\secondbest}[1]{\textcolor{blue}{\textbf{#1}}}

\usepackage{hyperref}

\usepackage{orcidlink}

\newcommand{\modelname}{LFNet}

\begin{document}

\title{Liquid Fusion of Heterogeneous Representations Towards General Salient Object Detection}

\titlerunning{LFNet}

\author{Ke Chen\inst{1}\orcidlink{0009-0005-6233-1731} \and
Ling Zhou\inst{2}\orcidlink{0000-0002-9508-4950} \and
Guangqi Jiang\inst{1}\orcidlink{0000-0001-9748-0407} \and
Gengshen Wu\inst{3}\orcidlink{0000-0003-3201-002X} \and
Yi Liu\inst{1}\thanks{Corresponding author.}\orcidlink{0000-0002-5177-5522} \and
Shoukun Xu\inst{1}\orcidlink{0000-0002-7119-1006}}

\authorrunning{K. Chen et al.}

\institute{School of Computer Science and Artificial Intelligence, Changzhou University, Changzhou, Jiangsu, 213159, China \and
College of Computer Science and Artificial Intelligence, Fudan University, Shanghai, 200082, China \and
Faculty of Data Science, City University of Macau, Avenida Padre Tomás Pereira Taipa, Macao, 999078, China\\
\email{s24150812007@smail.cczu.edu.cn, lzhou24@m.fudan.edu.cn, gswu@cityu.edu.mo, \{guangqijiang, liuyi0089, skxu\}@cczu.edu.cn}
}

\maketitle

\begin{abstract}
	General Salient Object Detection (SOD) aims to identify and segment visually interesting objects from uni-modality or multi-modality scenes, recently advanced by cutting-edge State Space Models (SSMs). However, a critical limitation of current approaches is their neglect of the inherent spectral biases exhibited by different neural network paradigms. By digging to the dataset-level spectral analysis of Convolutional Neural Networks (CNNs) and SSMs, their semantic representations are inherently complementary based on their complementary frequency preferences. Inspired by this, we harmonize heterogeneous representations from SSMs and CNNs to bridge their spectral biases for general salient object detection. To this end, inspired by the dynamic information propagation of Liquid Neural Networks (LNNs), we introduce a liquid fusion to dynamically integrates features from two backbones, including VMamba and ConvNeXt, referred to Liquid Fusion Network (\modelname). Concretely, by treating the continuous VMamba features and ConvNeXt features as evolving states and exogenous stimulus, respectively, \modelname\ employs a dynamic gating mechanism for content-aware feature aggregation. Crucially, this state-stimulus paradigm enables to scale to multi-modal cues, resulting in flexibility in general SOD. Besides, a Saliency-Guided Upsampling (SGU) operator to propagate the features to the shallow layer, which leverages a spectral-spatial co-design to suppress upsampling artifacts while preserving semantics. Extensive experiments across five diverse tasks (RGB, RGB-D, RGB-T, VSOD, and VDT) demonstrate that \modelname\ achieves state-of-the-art performance, offering a superior trade-off between detection accuracy and model efficiency. Code has been released at \url{https://github.com/cke520/LFNet}.
\end{abstract}

\section{Introduction}
\label{sec:intro}

Salient object detection, commonly referred to as SOD, is a fundamental computer vision task dedicated to identifying and precisely segmenting the most visually prominent objects within a scene. By effectively filtering out background clutter, this technique serves as an indispensable preliminary step for a multitude of downstream applications, including visual tracking~\cite{track}, image editing~\cite{edit}, scene understanding~\cite{scene,transcending}, autonomous driving~\cite{pwrf,driving}, and robotic navigation~\cite{tracking}.

A robust SOD model must construct comprehensive representations that satisfy two competing requirements: capturing broad semantic contexts for object localization and preserving fine-grained structural details for precise boundary delineation. Recently, the field has progressed toward \textbf{general SOD}~\cite{samba,vscode}, an ambitious framework encompassing single-modal RGB~\cite{icon,edn}, dual-modal RGB-D/T~\cite{swinnet,spnet} and Video SOD (VSOD)~\cite{ufo,ugpl}, as well as tri-modal Visible-Depth-Thermal (VDT) tasks~\cite{pwrf,dwfpr}. While Transformer-based methods~\cite{vscode,vst,vst++} established strong baselines, their quadratic complexity limits efficiency. Consequently, State Space Models~\cite{s4} (SSMs) like Mamba~\cite{mamba} have emerged as compelling alternatives. Notably, Samba~\cite{samba} successfully adapted SSMs to general SOD, achieving state-of-the-art results by constructing efficient long-range dependencies with linear computational complexity.

\begin{figure}[t]
	\centering
	\includegraphics[width=1\linewidth]{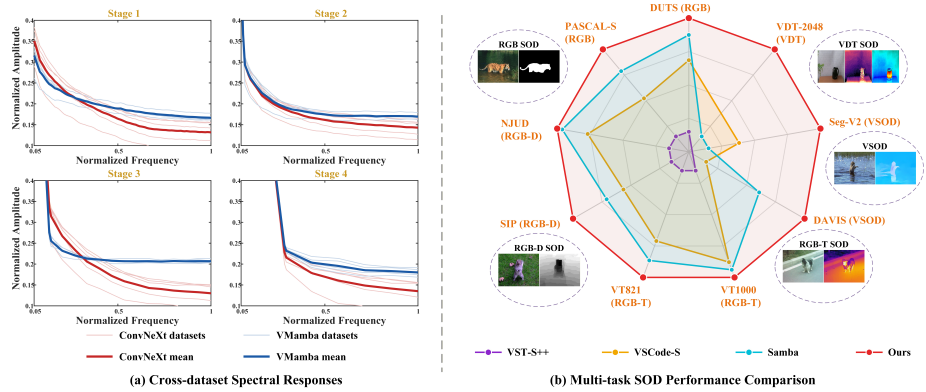} 
	\caption{\textbf{Spectral response and multi-task performance.} (a) Normalized frequency amplitude curves averaged over five representative datasets, i.e., PASCAL-S~\cite{pascals}, NJUD~\cite{njud}, VT5000~\cite{vt5000}, DAVIS~\cite{davis}, and VDT-2048~\cite{vdt2048}. The intersecting patterns show that ConvNeXt (red) and VMamba (blue) have different energy preferences across frequency bands, necessitating their fusion for balanced full-spectrum perception. (b) Performance comparison ($S_m$) across five general SOD tasks, demonstrating our superior performance.
	}
	\label{fig:motivation}
\end{figure}

From a signal processing perspective, the fundamental nature of SOD demands robust representations across the entire frequency spectrum. Current SSM-based approaches predominantly rely on engineering increasingly intricate spatial scanning strategies (e.g., cross-shaped or continuous paths) to force a single architecture to capture all necessary multi-scale contexts. However, we argue that this direction overlooks the inherent architectural biases dictated by different neural processing mechanisms. To investigate this, we conduct a cross-dataset spectral analysis over representative benchmarks covering RGB, RGB-D, RGB-T, VSOD, and VDT SOD. Specifically, we extract hierarchical features across four scaling stages from a lightweight Convolutional Neural Network~\cite{resnet} (ConvNeXt-Pico~\cite{convnext}) and an SSM network~\cite{s4} (VMamba-Small~\cite{vmamba}), applying 2D Fast Fourier Transform (FFT~\cite{fft}) to compute their radial frequency profiles. 

As explicitly demonstrated in Fig.~\ref{fig:motivation}~(a), the spectral energy distributions of CNNs and SSMs exhibit distinct divergence and intersection patterns across different scaling stages (e.g., stage 1 and stage 3). This observed structural heterogeneity reveals that CNNs and SSMs encode spatial information in fundamentally different ways, exhibiting distinct spectral signatures. This factual spectral divergence mathematically proves that their representations are inherently complementary, and relying solely on a single paradigm inevitably creates representational blind spots.

Inspired by this natural complementarity, we harmonize heterogeneous representations from SSMs and CNNs to bridge their spectral biases for general salient object detection. To this end, inspired by the continuous-time dynamics of Liquid Neural Networks (LNNs)~\cite{ltc,cfc}, we recast them to a liquid fusion that dynamically integrate features from two backbones, including VMamba and ConvNeXt, which is referred to Liquid Fusion Network (\modelname). Concretely, by treating the continuous VMamba features and ConvNeXt features as evolving states and exogenous stimulus, respectively, LFNet employs a dynamic gating mechanism for content-aware feature aggregation. Crucially, this state-stimulus paradigm enables to scale to multi-modal cues. This ensures flexibility in general SOD beyond RGB SOD, including RGB-D SOD, RGB-T SOD, VSOD, and VDT SOD. 

Besides, a Saliency-Guided Upsampling (SGU) operator to propagate the features to the shallow layer, which leverages a spectral-spatial co-design to suppress upsampling artifacts while preserving semantics. As evidenced by the quantitative comparisons in Fig.~\ref{fig:motivation}~(b), \modelname\ yields superior accuracy across diverse general SOD tasks.

In summary, our main contributions are four-fold:
\begin{itemize}
	\item We propose \modelname, a novel general SOD framework that harmonizes heterogeneous representations from SSMs and CNNs to bridge their inherent spectral biases.
	\item We design a liquid fusion based on a continuous state-stimulus mechanism to dynamically aggregate heterogeneous features, serving as a highly scalable solution for multi-modal fusion.
	\item We develop a saliency-guided upsampling operator, which utilizes a dual-branch spectral-spatial design to effectively propagate semantics during resolution restoration.
	\item Extensive experiments demonstrate that \modelname\ achieves state-of-the-art performance across five diverse general SOD tasks.
\end{itemize}
\section{Related Work}

\subsection{Salient Object Detection}

To address complex real-world scenarios, Salient Object Detection (SOD) has evolved from single-image processing to a diverse set of tasks, encompassing single-modal RGB, dual-modal RGB-D/T, Video SOD (VSOD), and tri-modal Visible-Depth-Thermal (VDT) SOD.

\textbf{RGB SOD.} Early RGB SOD research was predominantly driven by CNN-based architectures~\cite{basnet,csf-r2,edn,menet}, which struggle with long-range global dependency modeling due to limited receptive fields. While recent Transformer-based approaches~\cite{vst,vst++,vscode,icon} alleviate this by leveraging self-attention, their inherent quadratic computational complexity remains a critical bottleneck for high-resolution dense prediction tasks.

\textbf{RGB-D and RGB-T SOD.} To resolve visual ambiguities in challenging scenes, auxiliary modalities like depth (RGB-D~\cite{catnet,cpnet}) or thermal radiation (RGB-T~\cite{pcnet,contri}) are introduced. While effective, fusing these heterogeneous cues via intricate cross-modal attention~\cite{bbsnet} or dynamic guidance networks~\cite{spnet} often incurs significant computational overhead compared to rudimentary concatenation, highlighting the demand for more elegant integration mechanisms.

\textbf{Video SOD (VSOD).} VSOD extends detection into the temporal domain~\cite{ugpl,fsnet}, necessitating the joint modeling of spatial appearance and motion coherence. Existing approaches typically rely on precomputed optical flow or heavy spatiotemporal attention to capture inter-frame dependencies~\cite{mmnet,ufo,costformer}. Unfortunately, both paradigms suffer from high latency and computational redundancy, highlighting the critical need for highly efficient spatiotemporal modeling.

\textbf{VDT SOD.} Addressing extreme scenarios, VDT SOD leverages the synergy of visible, depth, and thermal modalities for robust detection. However, simultaneously aligning and fusing three heterogeneous inputs drastically increases model complexity~\cite{pwrf,mffnet}, typically sacrificing inference speed for accuracy. Ultimately, as the field progresses toward a general SOD framework, developing a scalable architecture that seamlessly handles single, dual, and tri-modal inputs—without suffering from quadratic computational costs—has become a pressing necessity.

\subsection{State Space Models}
Motivated by the success of sequence modeling in natural language processing, State Space Models (SSMs) have emerged as highly efficient alternatives to Transformers, achieving global receptive fields with linear computational complexity through hardware-aware selective scanning. Pioneering this shift, Vim~\cite{vim} successfully adapted bidirectional SSMs for visual tasks, while subsequent architectures like VMamba~\cite{vmamba} introduced refined visual state space blocks to enhance feature extraction across various domains, including semantic segmentation~\cite{sigma}, object detection~\cite{unetmamba}, and video understanding~\cite{hmba}. Within the specific domain of Salient Object Detection (SOD), Samba~\cite{samba} pioneered the integration of SSMs to construct efficient contexts. To overcome the inherent 1D nature of SSMs when processing 2D visual data, a predominant trend involves engineering increasingly intricate scanning strategies, such as multi-directional diagonal paths~\cite{rsmamba}, continuous 2D sequences~\cite{plainmamba}, or Nested S-shaped Scanning (NSS)~\cite{nss}. Despite these innovations, we argue that relying solely on increasingly complex scanning routes to force a single architecture to capture both broad semantics and fine-grained local textures often introduces significant architectural redundancy and optimization challenges. As revealed by our spectral analysis, pure SSMs suffer from inherent spectral biases that constrain their ability to represent high-frequency spatial features. This structural limitation highlights the fundamental difficulty of achieving comprehensive full-spectrum perception within a purely homogeneous SSM framework, thereby motivating our proposed heterogeneous hybrid paradigm that harmonizes SSMs with CNN-based local feature extraction.

\subsection{Liquid Neural Networks}
\label{subsec:liquid_neural_networks}
Liquid Neural Networks (LNNs) are continuous-time recurrent models governed by Liquid Time-Constant (LTC) dynamics~\cite{ltc}. Unlike traditional networks with fixed weights, LNNs offer highly expressive, input-dependent temporal adaptation~\cite{assess}. Initially bottlenecked by numerical ODE solvers, the practical application of LNNs was revitalized by the Closed-form Continuous-time (CfC) network~\cite{cfc}, which introduced a highly efficient analytical approximation. This breakthrough has propelled LNNs into diverse sequential and multimodal applications, including robotic control~\cite{robust} and edge computing~\cite{loihi}. In this paper, we recognize that the continuous-time dynamics of LNNs perfectly align with the challenge of fusing heterogeneous spatial features. By adapting the LNN philosophy into our proposed Liquid Fusion Module (LFM), we move beyond static element-wise addition, treating global SSM representations and local CNN features as an evolving dynamic system. To the best of our knowledge, this work represents the first attempt to leverage liquid dynamics for constructing a cross-paradigm heterogeneous fusion framework in dense visual prediction tasks, such as general SOD.

\section{Methodology}

\subsection{Preliminaries}

\textbf{Liquid Neural Networks and Closed-Form Dynamics. }
Unlike traditional recurrent models with fixed discrete dynamics, Liquid Neural Networks (LNNs)~\cite{ltc} exhibit continuous, input-dependent temporal behavior. The hidden state $\mathbf{x}(t)$ evolves under an exogenous input $\mathbf{I}(t)$ via a liquid time-constant (LTC) Ordinary Differential Equation (ODE):
\begin{equation}
	\frac{d\mathbf{x}(t)}{dt} = - \left[ w_\tau + f(\mathbf{x}, \mathbf{I}; \theta) \right] \odot \mathbf{x}(t) + A \odot f(\mathbf{x}, \mathbf{I}; \theta),
	\label{eq:ltc_ode}
\end{equation}
where $w_\tau$ represents the base time constant and $f(\cdot)$ denotes a nonlinearity. To bypass the computational bottleneck of stiff numerical solvers, Closed-form Continuous-depth (CfC) models~\cite{cfc} derive a bounded closed-form solution. Crucially, to mitigate vanishing gradients associated with exponential decay, CfC replaces the decay term with a learnable sigmoidal gate $\sigma(\cdot)$, effectively functioning as a dynamic interpolation mechanism. 

Inspired by this continuous-time gating mechanism~\cite{cfc}, we translate the temporal dynamics into the spatial domain. We formulate our Liquid Fusion dynamic as a weighted equilibrium between the evolving \textit{memory state} (derived from the continuous SSM stream $\mathbf{h}$) and the \textit{synaptic stimulus} (derived from the static CNN stream $\tilde{\mathbf{l}}$), defined as:
\begin{equation}
	\mathbf{x}_{out} = (1 - \sigma) \odot \mathbf{h} + \sigma \odot \tilde{\mathbf{l}},
	\label{eq:liquid_fusion_theory}
\end{equation}
where the dynamic permeability $\sigma \in (0, 1)$ regulates the injection of exogenous stimuli. In this context, high permeability ($\sigma \to 1$) facilitates rapid stimulus integration, while low permeability ($\sigma \to 0$) prioritizes the preservation of the inherent memory state.

\begin{figure}[t]
	\centering
	\includegraphics[width=0.92\textwidth]{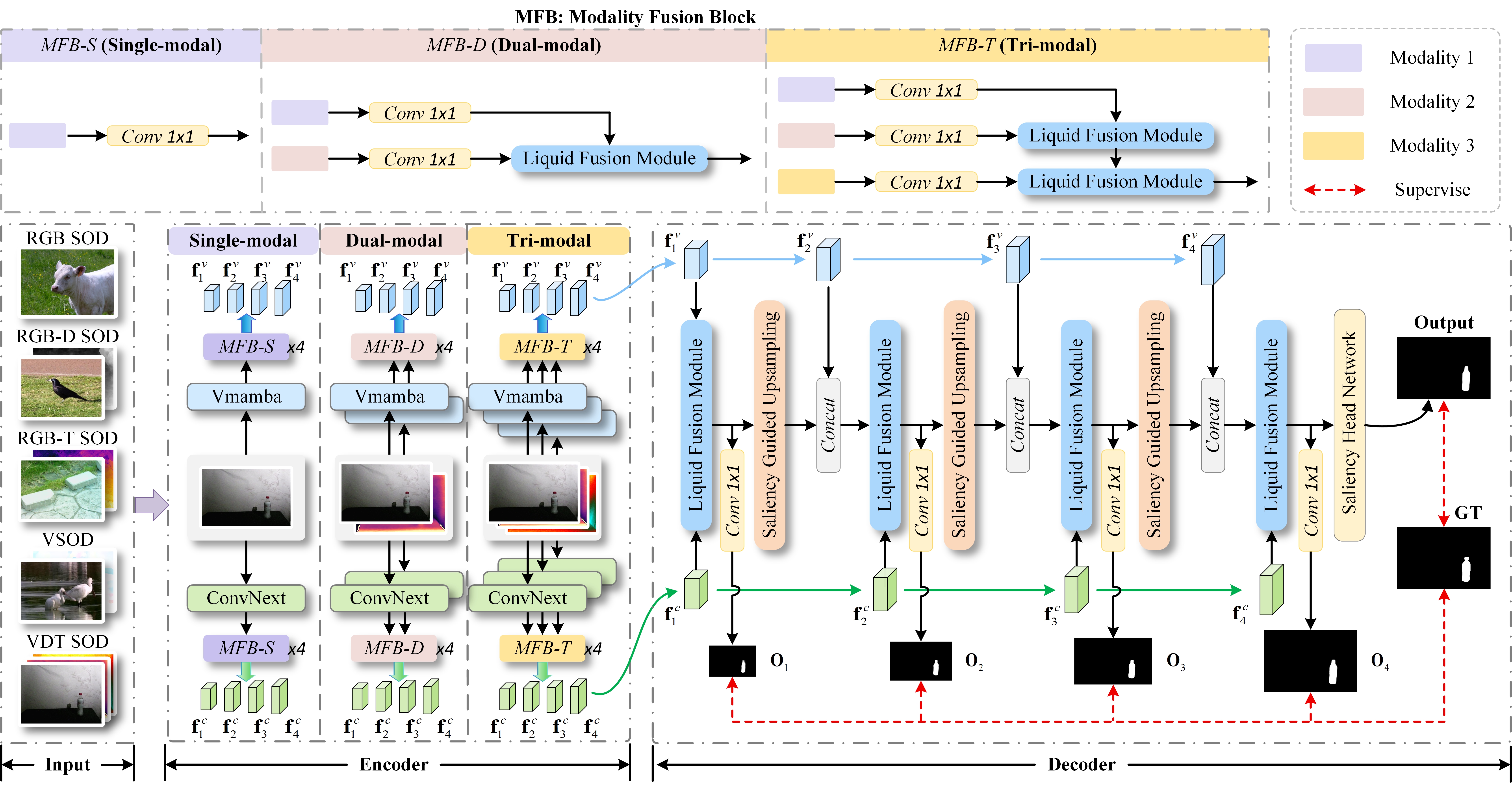}
	\caption{\textbf{Overall architecture of the proposed \modelname.} Given varying input modalities, parallel VMamba and ConvNeXt backbones extract heterogeneous features, which are then harmonized by Modality Fusion Blocks (MFB) and progressively aggregated through a top-down decoder using Liquid Fusion Modules (LFMs) and Saliency-Guided Upsampling (SGU) operators with multi-scale supervision.}
	\label{fig:overview}
\end{figure}

\subsection{Overview}
Inspired by the distinct yet complementary frequency preferences observed between VMamba and ConvNeXt, we propose \modelname\ to harmonize these heterogeneous representations for general SOD. As illustrated in Fig.~\ref{fig:overview}, our Heterogeneous Hybrid Encoder extracts representations across four hierarchical stages: a VMamba stream captures continuous sequence-based semantics, while a parallel ConvNeXt stream preserves grid-based spatial inductive biases. To seamlessly integrate these heterogeneous cues, our decoder employs the novel Liquid Fusion Module (LFM), an LNN-inspired gated fusion module, for content-aware dynamic feature aggregation. Finally, the Saliency-Guided Upsampling (SGU) module progressively reconstructs high-resolution predictions, utilizing a spectral-spatial co-design to propagate semantics to shallow layers during resolution restoration.

\subsection{Heterogeneous Hybrid Encoder and Modality Fusion}
To leverage the structural heterogeneity revealed in our spectral analysis, we construct a Heterogeneous Hybrid Encoder integrating two distinct backbones. We employ a pre-trained VMamba-Small~\cite{vmamba} to generate continuous state-space representations with linear complexity, yielding features $\mathbf{f}^v_i \in \mathbb{R}^{C_i \times H_i \times W_i}$ at stages $i \in \{1, 2, 3, 4\}$. Complementing this, a parallel ConvNeXt-Pico~\cite{convnext} stream introduces a distinct architectural inductive bias, ensuring the capture of comprehensive spectral cues without relying on complex scanning engineering. To handle diverse inputs, a Modality Fusion Block (\textit{MFB}) within the ConvNeXt stream generates task-adapted features $\mathbf{f}^c_i$. The MFB applies a $1\times1$ projection for single-modal RGB inputs, uses a single LFM for dual-modal tasks (RGB-D/T, VSOD), and employs a cascaded LFM structure for tri-modal VDT scenarios, effectively absorbing auxiliary cues without the quadratic cost of cross-attention.

\begin{figure}[t]
	\centering
	\includegraphics[width=0.8\columnwidth]{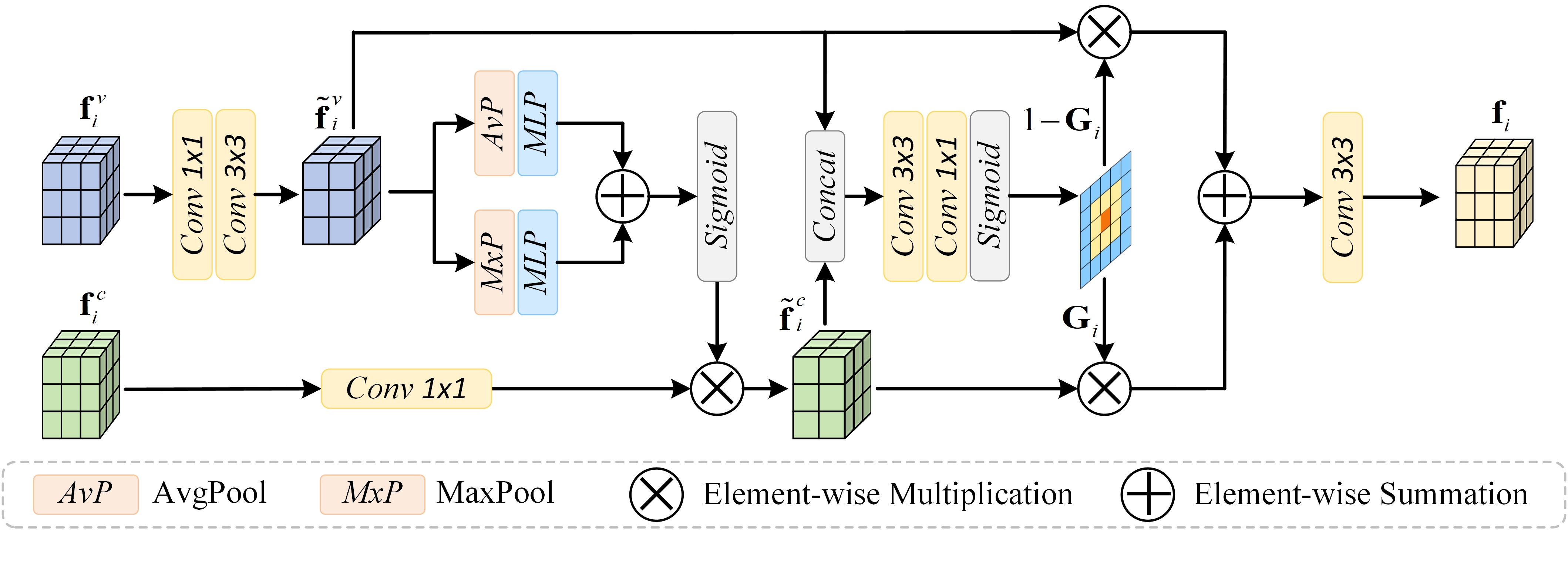} 
	\caption{Detailed architecture of the Liquid Fusion Module (LFM).}
	\label{fig:lfm}
\end{figure}

\subsection{Liquid Fusion Module}

Inspired by the dynamic equilibrium of Liquid Neural Networks~\cite{cfc}, our LFM (detailed in Fig.~\ref{fig:lfm}) reformulates heterogeneous feature integration as a \textit{state-stimulus} interaction process. We treat the VMamba feature $\mathbf{f}^v_i$ as an evolving \textit{memory state} and the ConvNeXt feature $\mathbf{f}^c_i$ as an \textit{exogenous stimulus}. Initially, the VMamba state is projected into an embedding $\tilde{\mathbf{f}}^v_i$ via sequential $1\times1$ and $3\times3$ convolutions.

\noindent\textbf{Adaptive Channel Modulation. } To adaptively filter task-irrelevant details, we modulate the ConvNeXt stimulus using a channel attention map derived from the VMamba state. This modulation is mathematically expressed as:
\begin{equation}
	\tilde{\mathbf{f}}^c_i = \textit{Conv}_{1\times1}(\mathbf{f}^c_i) \odot \sigma \left( \mathcal{M}(\textit{AvgP}(\tilde{\mathbf{f}}^v_i)) + \mathcal{M}(\textit{MaxP}(\tilde{\mathbf{f}}^v_i)) \right),
\end{equation}
where $\mathcal{M}$ denotes a shared Multi-Layer Perceptron (MLP), $\textit{AvgP}$ and $\textit{MaxP}$ represent spatial average and max pooling operations respectively, and $\sigma$ designates the Sigmoid activation function. 

\noindent\textbf{Dynamic Spatial Gating. } To achieve a dynamic balance between memory retention and stimulus injection, we first compute a spatial liquid gate $\mathbf{G}_i$ based on the concatenated features $[\tilde{\mathbf{f}}^v_i, \tilde{\mathbf{f}}^c_i]$, calculated as:
\begin{equation}
	\mathbf{G}_i = \sigma \left( \textit{Conv}_{1\times1} \left( \textit{Conv}_{3\times3} ( [\tilde{\mathbf{f}}^v_i, \tilde{\mathbf{f}}^c_i] ) \right) \right).
\end{equation}
Leveraging the theoretical foundation established in Eq.~\eqref{eq:liquid_fusion_theory}, we materialize the fusion process as a closed-form dynamic selection mechanism:
\begin{equation}
	\mathbf{f}_i = \textit{Conv}_{3\times3} \left( (1 - \mathbf{G}_i) \odot \tilde{\mathbf{f}}^v_i + \mathbf{G}_i \odot \tilde{\mathbf{f}}^c_i \right).
\end{equation}
Here, $\mathbf{G}_i$ serves as the dynamic permeability parameter (analogous to $\sigma$ in Eq.~\eqref{eq:liquid_fusion_theory}). Consequently, as $\mathbf{G}_i \to 1$, the network prioritizes the injection of grid-based spatial cues from ConvNeXt (stimulus); conversely, as $\mathbf{G}_i \to 0$, it prioritizes retaining the sequence-based context from VMamba (memory). This adaptive mechanism effectively reconciles the distinct architectural biases of the two streams.

\begin{figure}[t]
	\centering
	\includegraphics[width=0.75\columnwidth]{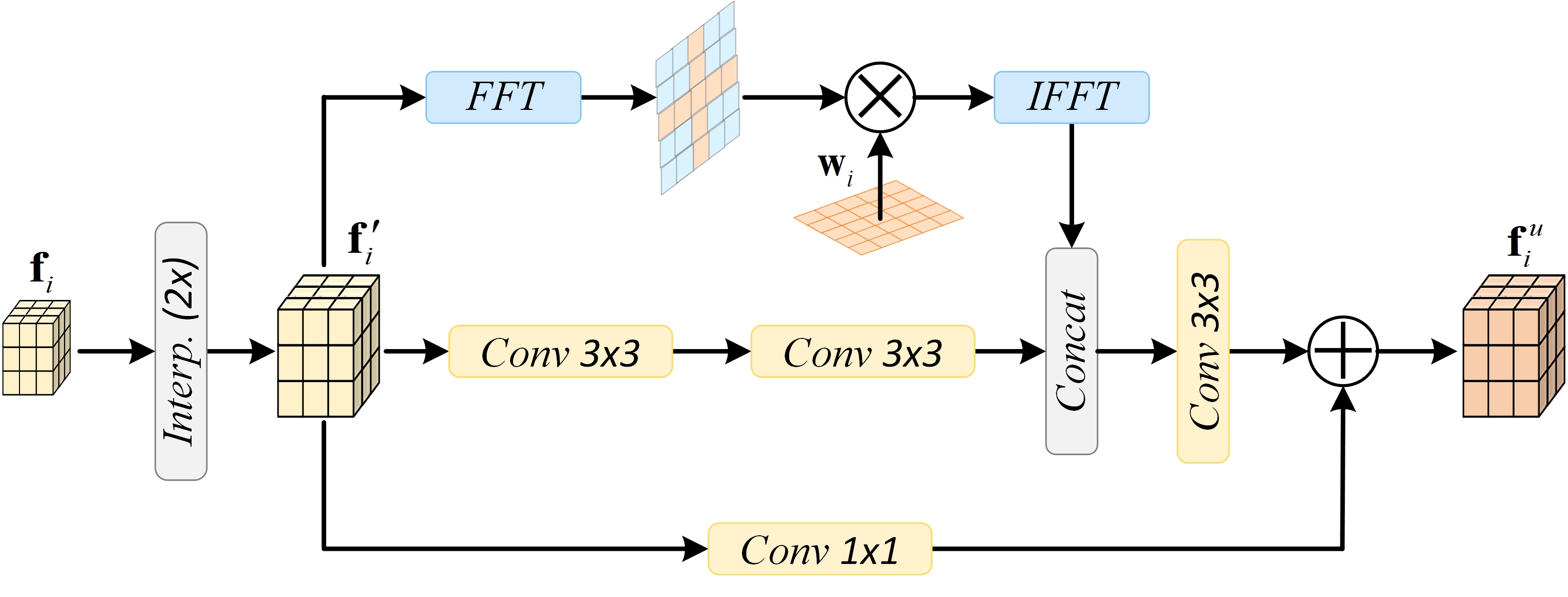}
	\caption{Detailed architecture of the Saliency-Guided Upsampling (SGU).}
	\label{fig:sgu}
\end{figure}

\subsection{Saliency-Guided Upsampling (SGU)}

Standard bilinear upsampling often leads to spectral aliasing and blurred object boundaries. To preserve sharp details for SOD, our SGU module (illustrated in Fig.~\ref{fig:sgu}) employs a dual-branch design to enforce consistency in both spatial and frequency domains.

\noindent\textbf{Spectral-Spatial Co-Design. } We first upscale the fused feature $\mathbf{f}_i$ via $2\times$ interpolation to yield the intermediate feature $\mathbf{f}'_i$. To preserve global shape integrity, the spectral branch transforms $\mathbf{f}'_i$ via Fast Fourier Transform (FFT) and modulates the spectrum with a learnable complex weight matrix $\mathbf{w}_i$, formulated as:
\begin{equation}
	\mathbf{F}^{spec}_i = \textit{IFFT} \left( \textit{FFT}(\mathbf{f}'_i) \odot \mathbf{w}_i \right),
\end{equation}
where $\textit{IFFT}$ denotes the Inverse FFT. Concurrently, the spatial branch captures high-frequency edge cues via two stacked $3\times3$ convolutions to generate the spatial representation $\mathbf{F}^{spat}_i$.

\noindent\textbf{Dual-Domain Fusion. } The final reconstructed feature $\mathbf{f}^{u}_i$ integrates both domains alongside a residual connection, derived as:
\begin{equation}
	\mathbf{f}^{u}_i = \textit{Conv}_{3\times3}\left( [\mathbf{F}^{spec}_i, \mathbf{F}^{spat}_i] \right) + \textit{Conv}_{1\times1}(\mathbf{f}'_i).
\end{equation}
By synergizing spatial precision with spectral consistency, SGU ensures that the decoding phase restores sharp boundaries without suffering from structural degradation.

\subsection{Loss Function}
To facilitate robust feature learning, we employ a deep supervision strategy. The total loss $\mathcal{L}_{total}$ is defined as a weighted combination of the Binary Cross Entropy (BCE) loss and the Intersection-over-Union (IoU) loss, applied to predictions at all scales:
\begin{equation}
	\mathcal{L}_{total} = \sum_{k=1}^{4} \left( \mathcal{L}_{bce}(\mathbf{O}_k, \mathbf{GT}) + \mathcal{L}_{iou}(\mathbf{O}_k, \mathbf{GT}) \right),
\end{equation}
where $\mathbf{O}_k$ represents the prediction map at stage $k$ and $\mathbf{GT}$ denotes the ground truth. This multi-scale supervision ensures that the model learns discriminative features progressively from coarse to fine levels.

\section{Experiments and Results}

\subsection{Datasets and Metrics}
To comprehensively evaluate the proposed \modelname, we conduct extensive experiments across five distinct tasks: RGB SOD, RGB-D SOD, RGB-T SOD, VSOD, and VDT SOD. 
For the RGB SOD task, we utilize five standard benchmark datasets: DUTS~\cite{duts}, DUT-O~\cite{duto}, HKU-IS~\cite{hkuis}, PASCAL-S~\cite{pascals}, and ECSSD~\cite{ecssd}. 
For RGB-D SOD, the evaluation is performed on NJUD~\cite{njud}, NLPR~\cite{nlpr}, SIP~\cite{sip}, STERE~\cite{stere}, and DUTLF-D~\cite{dutlfd}. 
Regarding RGB-T SOD, three benchmarks are employed: VT821~\cite{vt821}, VT1000~\cite{vt1000}, and VT5000~\cite{vt5000}. 
For VSOD, we use DAVIS~\cite{davis}, DAVSOD-easy~\cite{davsod}, FBMS~\cite{fbms}, SegV2~\cite{segv2}, and VOS~\cite{vos}. 
Lastly, for VDT SOD, we employ the VDT-2048~\cite{vdt2048} benchmark dataset.

To quantitatively assess the model performance, we adopt three widely used saliency metrics: structure-measure ($S_m$), maximum F-measure ($F_m$), and maximum enhanced-alignment measure ($E_m$). Furthermore, to evaluate the model size and computational complexity, we report the total number of parameters (Params).

\subsection{Implementation Details}
The proposed \modelname\ is implemented in PyTorch and trained on a single NVIDIA RTX 4090 D GPU. Following standard protocols, we utilize established training splits for all five tasks, rigorously removing duplicate samples in joint datasets to prevent data leakage. During training, all input images are resized to $512 \times 512$ and augmented using random flipping, cropping, and rotation. The model is optimized for 50 epochs with a batch size of 2 using the AdamW optimizer. The base learning rate is initialized at $1 \times 10^{-4}$ with a weight decay of 0.05, while the pre-trained backbone networks use only five percent of this rate to preserve representational integrity. We apply a five-epoch linear warmup followed by a cosine annealing decay schedule. To enhance training efficiency and prevent gradient explosion, Automatic Mixed Precision (AMP) and gradient clipping (with a maximum norm threshold of 0.5) are integrated. The optimal model weights are selected based on the highest S-measure achieved on the validation set.

\begin{table}[t]
	\centering
	\caption{Quantitative comparison of our \modelname\ against other SOTA RGB SOD methods on five benchmark datasets. ``-'' indicates the result is not available. ``$\uparrow$'' denotes larger is better. All metric values are in \%. The best two results are highlighted in \best{red} and \secondbest{blue}.}
	\label{tab:main_results}
	\resizebox{\textwidth}{!}{
		\setlength{\tabcolsep}{2.5pt} 
		\renewcommand{\arraystretch}{1.1}
		\begin{tabular}{l | c | ccc | ccc | ccc | ccc | ccc}
			\toprule
			\multirow{2}{*}{Method} & Params & \multicolumn{3}{c|}{DUTS~\cite{duts}} & \multicolumn{3}{c|}{DUT-O~\cite{duto}} & \multicolumn{3}{c|}{HKU-IS~\cite{hkuis}} & \multicolumn{3}{c|}{PASCAL-S~\cite{pascals}} & \multicolumn{3}{c}{ECSSD~\cite{ecssd}} \\
			\cmidrule(lr){3-5} \cmidrule(lr){6-8} \cmidrule(lr){9-11} \cmidrule(lr){12-14} \cmidrule(lr){15-17} 
			& (M) & \(S_m \uparrow\) & \(F_m \uparrow\) & \(E_m \uparrow\) & \(S_m \uparrow\) & \(F_m \uparrow\) & \(E_m \uparrow\) & \(S_m \uparrow\) & \(F_m \uparrow\) & \(E_m \uparrow\) & \(S_m \uparrow\) & \(F_m \uparrow\) & \(E_m \uparrow\) & \(S_m \uparrow\) & \(F_m \uparrow\) & \(E_m \uparrow\) \\
			\midrule
			\rowcolor{gray!10}
			\multicolumn{17}{c}{\textit{\textbf{CNN-based Methods}}} \\
			\midrule
			CSF-R2~\cite{csf-r2}    & 36.53 & 89.0 & 86.9 & 92.9 & 83.8 & 77.5 & 86.9 & -    & -    & -    & 86.3 & 83.9 & 88.5 & 93.1 & 94.2 & 96.0 \\
			EDN~\cite{edn}       & 42.85 & 89.2 & 89.3 & 93.3 & 84.9 & 82.1 & 88.4 & 92.4 & 94.0 & 96.3 & 86.4 & 87.9 & 90.7 & 92.7 & 95.0 & 95.7 \\
			ICON-R~\cite{icon}    & 33.09 & 89.0 & 87.6 & 93.1 & 84.5 & 79.9 & 88.4 & 92.0 & 93.1 & 96.0 & 86.2 & 84.4 & 88.8 & 92.8 & 94.3 & 96.0 \\
			MENet~\cite{menet}     & 27.83 & 90.5 & 89.5 & 94.3 & 85.0 & 79.2 & 87.9 & 92.7 & 93.9 & 96.5 & 87.1 & 84.8 & 89.2 & 92.7 & 93.8 & 95.6 \\
			\midrule
			\rowcolor{gray!10}
			\multicolumn{17}{c}{\textit{\textbf{Transformer-based Methods}}} \\
			\midrule
			ICON-S~\cite{icon}    & 94.30 & 91.7 & 91.1 & 96.0 & 86.9 & 83.0 & 90.6 & 93.6 & 94.7 & 97.4 & 88.5 & 86.0 & 90.3 & 94.1 & 95.4 & 97.1 \\
			BBRF~\cite{bbrf}      & 74.40 & 90.8 & 90.5 & 95.1 & 85.5 & 82.0 & 89.8 & 93.5 & 94.6 & 93.6 & 87.1 & 88.4 & 92.5 & 93.9 & 95.7 & 97.2 \\
			VST-S ++~\cite{vst++}  & 74.90 & 90.9 & 89.7 & 94.7 & 85.9 & 81.3 & 89.0 & 93.2 & 94.1 & 96.9 & 88.0 & 85.9 & 90.1 & 93.9 & 95.1 & 96.9 \\
			VSCode-S~\cite{vscode}  & 74.72 & 92.6 & 92.2 & 96.0 & 87.7 & 84.0 & 91.2 & 94.0 & 95.1 & 97.4 & 88.7 & 86.4 & 90.4 & 94.9 & 95.9 & 97.4 \\
			\midrule
			\rowcolor{gray!10}
			\multicolumn{17}{c}{\textit{\textbf{Mamba-based Methods}}} \\
			\midrule
			\rowcolor{orange!10}
			Samba~\cite{samba}     & 49.59 & \secondbest{93.2} & \secondbest{93.0} & \secondbest{96.6} & \secondbest{88.9} & \secondbest{85.9} & \secondbest{92.2} & \secondbest{94.5} & \secondbest{95.6} & \secondbest{97.8} & \secondbest{89.2} & \secondbest{89.6} & \secondbest{93.1} & \secondbest{95.3} & \secondbest{96.5} & \best{97.8} \\
			\rowcolor{orange!20}
			\textbf{Ours} & 43.23 & \best{93.6} & \best{93.6} & \best{97.0} & \best{90.4} & \best{88.4} & \best{93.8} & \best{94.6} & \best{95.8} & \best{97.9} & \best{89.6} & \best{89.8} & \best{93.9} & \best{95.4} & \best{96.7} & \secondbest{97.7} \\
			\bottomrule
		\end{tabular}
	}
\end{table}

\begin{table}[t]
	\centering
	\caption{Quantitative comparison of our \modelname\ against other SOTA RGB-D SOD methods on five benchmark datasets.}
	\label{tab:rgbd_results}
	\resizebox{\textwidth}{!}{
		\setlength{\tabcolsep}{2.5pt}
		\renewcommand{\arraystretch}{1.1}
		\begin{tabular}{l | c | ccc | ccc | ccc | ccc | ccc}
			\toprule
			\multirow{2}{*}{Method} & Params & \multicolumn{3}{c|}{NJUD~\cite{njud}} & \multicolumn{3}{c|}{NLPR~\cite{nlpr}} & \multicolumn{3}{c|}{SIP~\cite{sip}} & \multicolumn{3}{c|}{STERE~\cite{stere}} & \multicolumn{3}{c}{DUTLF-D~\cite{dutlfd}} \\
			\cmidrule(lr){3-5} \cmidrule(lr){6-8} \cmidrule(lr){9-11} \cmidrule(lr){12-14} \cmidrule(lr){15-17} 
			& (M) & \(S_m \uparrow\) & \(F_m \uparrow\) & \(E_m \uparrow\) & \(S_m \uparrow\) & \(F_m \uparrow\) & \(E_m \uparrow\) & \(S_m \uparrow\) & \(F_m \uparrow\) & \(E_m \uparrow\) & \(S_m \uparrow\) & \(F_m \uparrow\) & \(E_m \uparrow\) & \(S_m \uparrow\) & \(F_m \uparrow\) & \(E_m \uparrow\) \\
			\midrule
			\rowcolor{gray!10}
			\multicolumn{17}{c}{\textit{\textbf{CNN-based Methods}}} \\
			\midrule
			JL-DCF~\cite{jl-dcf}  & 143.52 & 87.7 & 89.2 & 94.1 & 93.1 & 91.8 & 96.5 & 88.5 & 89.4 & 93.1 & 90.0 & 89.5 & 94.2 & 89.4 & 89.1 & 92.7 \\
			SP-Net~\cite{sp-net}  & 67.88  & 92.5 & 92.8 & 95.7 & 92.7 & 91.9 & 96.2 & 89.4 & 90.4 & 93.3 & 90.7 & 90.6 & 94.9 & 89.5 & 89.9 & 93.3 \\
			DCF~\cite{dcf}     & 53.92  & 90.4 & 90.5 & 94.3 & 92.2 & 91.0 & 95.7 & 87.4 & 88.6 & 92.2 & 90.6 & 90.4 & 94.8 & 92.5 & 93.0 & 95.6 \\
			SPSN~\cite{spsn}    & -  & 91.8 & 92.1 & 95.2 & 92.3 & 91.2 & 96.0 & 89.2 & 90.0 & 93.6 & 90.7 & 90.2 & 94.5 & - & - & - \\
			\midrule
			\rowcolor{gray!10}
			\multicolumn{17}{c}{\textit{\textbf{Transformer-based Methods}}} \\
			\midrule
			CATNet~\cite{catnet} & 262.73 & 93.2 & 93.7 & 96.0 & 93.8 & 93.4 & 97.1 & 91.0 & 92.8 & 95.1 & 92.0 & 92.2 & 95.8 & 95.2 & 95.8 & 97.5 \\
			VST-S ++~\cite{vst++}  & 143.15 & 92.8 & 92.8 & 95.7 & 93.5 & 92.5 & 96.4 & 90.4 & 91.8 & 94.6 & 92.1 & 91.6 & 95.4 & 94.5 & 95.0 & 96.9 \\
			CPNet~\cite{cpnet} & 216.50 & 93.5 & 94.1 & 96.3 & 94.0 & 93.6 & 97.1 & 90.7 & 92.7 & 94.6 & 92.0 & 92.2 & 96.0 & 95.1 & 95.9 & 97.4 \\
			VSCode-S~\cite{vscode}  & 74.72 & 94.4 & 94.9 & 97.0 & 94.1 & 93.2 & 96.8 & 92.4 & 94.2 & 95.8 & 93.1 & 92.8 & 95.8 & \secondbest{96.0} & \secondbest{96.7} & \secondbest{98.0} \\
			\midrule
			\rowcolor{gray!10}
			\multicolumn{17}{c}{\textit{\textbf{Mamba-based Methods}}} \\
			\midrule
			\rowcolor{orange!10}
			Samba~\cite{samba}     & 54.94 & \secondbest{94.9} & \secondbest{95.6} & \secondbest{97.5} & \secondbest{94.7} & \secondbest{94.1} & \secondbest{97.6} & \secondbest{93.1} & \secondbest{94.9} & \secondbest{96.6} & \secondbest{93.5} & \secondbest{93.3} & \secondbest{96.3} & 95.6 & 96.4 & 97.6 \\
			\rowcolor{orange!20}
			\textbf{Ours} & {44.65} & \best{95.0} & \best{95.8} & \best{97.8} & \best{94.9} & \best{94.3} & \best{97.7} & \best{94.5} & \best{96.1} & \best{97.7} & \best{93.6} & \best{93.5} & \best{96.4} & \best{96.3} & \best{97.3} & \best{98.4} \\
			\bottomrule
		\end{tabular}
	}
\end{table}

\begin{table}[t]
	\centering
	\caption{Quantitative comparison of our \modelname\ against other SOTA RGB-T SOD methods on three benchmark datasets.}
	\label{tab:rgbt_results}
	\resizebox{\textwidth}{!}{
		\setlength{\tabcolsep}{2.5pt}
		\renewcommand{\arraystretch}{0.8}
		\begin{tabular}{l | c | cccc | cccc | >{\columncolor{orange!10}}c >{\columncolor{orange!20}}c}
			\toprule
			\multicolumn{2}{c|}{\multirow{2}{*}{Method}} & \multicolumn{4}{c|}{\cellcolor{gray!10}\textit{\textbf{CNN-based}}} & \multicolumn{4}{c|}{\cellcolor{gray!10}\textit{\textbf{Transformer-based}}} & \multicolumn{2}{c}{\cellcolor{gray!10}\textit{\textbf{Mamba-based}}} \\
			\cmidrule(lr){3-6} \cmidrule(lr){7-10} \cmidrule(lr){11-12}
			\multicolumn{2}{c|}{} & \makecell{MGAI \\ ~\cite{mgai}} & \makecell{TNet \\ ~\cite{tnet}} & \makecell{CGMDR \\ ~\cite{cgmdr}} & \makecell{SPNet \\ ~\cite{spnet}} & \makecell{VST-S ++ \\ ~\cite{vst++}} & \makecell{VSCode-S \\ ~\cite{vscode}} & \makecell{PCNet \\ ~\cite{pcnet}} & \makecell{ConTri \\ ~\cite{contri}} & \makecell{Samba \\ ~\cite{samba}} & \textbf{Ours} \\
			\midrule
			\multicolumn{2}{c|}{Params (M)} & 87.09 & 87.04 & - & 104.03 & 143.15 & 74.72 & - & 96.31 & 54.94 & 44.65 \\
			\midrule
			\multirow{3}{*}{\makecell{VT821 \\ \cite{vt821}}}  & \(S_m \uparrow\) & 89.1 & 89.9 & 89.4 & 91.3 & 89.7 & 92.6 & 91.5 & 91.6 & \secondbest{93.4} & \best{94.1} \\
			& \(F_m \uparrow\) & 87.0 & 88.5 & 87.2 & 90.0 & 86.8 & 91.0 & 87.9 & 89.6 & \secondbest{92.7} & \best{93.2} \\
			& \(E_m \uparrow\) & 93.3 & 93.6 & 93.2 & 94.9 & 92.5 & 95.4 & 94.1 & 94.8 & \secondbest{96.5} & \best{96.8} \\
			\midrule
			\multirow{3}{*}{\makecell{VT1000 \\ \cite{vt1000}}} & \(S_m \uparrow\) & 92.9 & 92.9 & 93.1 & 94.1 & 94.0 & 95.2 & 94.3 & 94.1 & \secondbest{95.3} & \best{95.4} \\
			& \(F_m \uparrow\) & 92.1 & 92.1 & 92.7 & 94.3 & 93.1 & \secondbest{94.7} & 92.4 & 93.9 & \best{95.6} & \best{95.6} \\
			& \(E_m \uparrow\) & 96.5 & 96.5 & 96.6 & 97.5 & 97.1 & 98.1 & 95.8 & 97.6 & \best{98.3} & \secondbest{98.2} \\
			\midrule
			\multirow{3}{*}{\makecell{VT5000 \\ \cite{vt5000}}} & \(S_m \uparrow\) & 88.4 & 89.5 & 89.6 & 91.4 & 90.1 & \secondbest{92.5} & 92.0 & 92.4 & \best{92.8} & \best{92.8} \\
			& \(F_m \uparrow\) & 84.6 & 86.4 & 87.7 & 90.5 & 86.1 & 90.0 & 89.9 & 91.8 & \secondbest{91.9} & \best{92.0} \\
			& \(E_m \uparrow\) & 93.0 & 93.6 & 93.9 & 95.4 & 93.6 & 95.9 & 95.6 & \secondbest{96.3} & \best{96.3} & \best{96.3} \\
			\bottomrule
		\end{tabular}
	}
\end{table}

\begin{table}[t]
	\centering
	\caption{Quantitative comparison of our \modelname\ against other SOTA VSOD methods on five benchmark datasets.}
	\label{tab:vsod_results}
	\resizebox{\textwidth}{!}{
		\setlength{\tabcolsep}{2.5pt}
		\renewcommand{\arraystretch}{1.1}
		\begin{tabular}{l | c | ccc | ccc | ccc | ccc | ccc}
			\toprule
			\multirow{2}{*}{Method} & Params & \multicolumn{3}{c|}{DAVIS~\cite{davis}} & \multicolumn{3}{c|}{DAVSOD-easy~\cite{davsod}} & \multicolumn{3}{c|}{FBMS~\cite{fbms}} & \multicolumn{3}{c|}{SegV2~\cite{segv2}} & \multicolumn{3}{c}{VOS~\cite{vos}} \\
			\cmidrule(lr){3-5} \cmidrule(lr){6-8} \cmidrule(lr){9-11} \cmidrule(lr){12-14} \cmidrule(lr){15-17} 
			& (M) & \(S_m \uparrow\) & \(F_m \uparrow\) & \(E_m \uparrow\) & \(S_m \uparrow\) & \(F_m \uparrow\) & \(E_m \uparrow\) & \(S_m \uparrow\) & \(F_m \uparrow\) & \(E_m \uparrow\) & \(S_m \uparrow\) & \(F_m \uparrow\) & \(E_m \uparrow\) & \(S_m \uparrow\) & \(F_m \uparrow\) & \(E_m \uparrow\) \\
			\midrule
			\rowcolor{gray!10}
			\multicolumn{17}{c}{\textit{\textbf{CNN-based Methods}}} \\
			\midrule
			FSNet~\cite{fsnet} & 83.41 & 92.2 & 90.9 & 97.2 & 76.0 & 63.7 & 79.6 & 87.5 & 86.7 & 91.8 & 84.9 & 77.3 & 92.0 & 67.8 & 62.1 & 75.5 \\
			DCFNet~\cite{dcfnet} & 69.56 & 91.4 & 89.9 & 97.0 & 72.9 & 61.2 & 78.1 & 88.3 & 85.3 & 91.0 & 90.3 & 87.0 & 95.3 & 83.8 & 77.3 & 86.1 \\
			UGPL~\cite{ugpl} & -  & 91.1 & 89.5 & 96.8 & 73.2 & 60.2 & 77.1 & 89.7 & 88.4 & 93.9 & 86.7 & 82.8 & 93.8 & 75.1 & 68.5 & 81.1 \\
			MMNet~\cite{mmnet} & 50.81 & 89.6 & 87.7 & 96.3 & 74.8 & 63.6 & 79.7 & 88.3 & 86.5 & 92.1 & 90.3 & 88.3 & 95.5 & - & - & - \\
			\midrule
			\rowcolor{gray!10}
			\multicolumn{17}{c}{\textit{\textbf{Transformer-based Methods}}} \\
			\midrule
			MGTNet~\cite{mgtnet}    & 150.91 & 92.5 & 91.9 & 97.6 & 76.5 & 65.3 & 80.0 & 90.0 & 88.1 & 92.9 & 90.3 & 86.1 & 94.6 & 81.4 & 72.7 & 81.9 \\
			UFO~\cite{ufo}       & 55.92 & 91.8 & 90.6 & 97.8 & 74.7 & 62.6 & 79.9 & 85.8 & 86.8 & 91.1 & 88.8 & 85.0 & 95.1 & - & - & - \\
			CoSTFormer~\cite{costformer} & - & 92.3 & 90.6 & 97.8 & 77.9 & 66.7 & 81.9 & 86.9 & 86.1 & 91.3 & 87.4 & 81.3 & 94.3 & 79.1 & 70.8 & 81.1 \\
			VSCode-S~\cite{vscode} & 74.72 & 93.6 & 92.2 & 97.3 & 80.0 & 71.0 & 83.5 & 90.5 & 90.2 & 93.9 & \secondbest{94.6} & 93.7 & 98.4 & - & - & - \\
			\midrule
			\rowcolor{gray!10}
			\multicolumn{17}{c}{\textit{\textbf{Mamba-based Methods}}} \\
			\midrule
			\rowcolor{orange!10}
			Samba~\cite{samba}     & 54.94 & \secondbest{94.3} & \secondbest{93.6} & \secondbest{98.5} & \secondbest{81.3} & \secondbest{73.4} & \secondbest{85.6} & \secondbest{92.5} & \secondbest{92.2} & \secondbest{95.4} & 94.3 & \secondbest{93.8} & \secondbest{98.7} & \secondbest{87.0} & \secondbest{82.0} & \secondbest{89.8} \\
			\rowcolor{orange!20}
			\textbf{Ours} & 44.65 & \best{94.9} & \best{94.4} & \best{98.8} & \best{83.1} & \best{76.2} & \best{87.5} & \best{92.7} & \best{92.9} & \best{96.4} & \best{95.4} & \best{95.0} & \best{99.2} & \best{88.9} & \best{84.4} & \best{91.0} \\
			\bottomrule
		\end{tabular}
	}
\end{table}

\begin{table}[t]
	\centering
	\caption{Quantitative comparison of our \modelname\ against other SOTA methods on the VDT-2048 benchmark dataset.}
	\label{tab:vdt_results_comparison}
	\resizebox{\textwidth}{!}{
		\setlength{\tabcolsep}{2.5pt}
		\renewcommand{\arraystretch}{0.8}
		\begin{tabular}{l | c | cc | cc | cccc | >{\columncolor{orange!10}}c >{\columncolor{orange!20}}c}
			\toprule
			\multicolumn{2}{c|}{\multirow{2}{*}{Method}} & \multicolumn{2}{c|}{\cellcolor{gray!10}\textit{\textbf{RGB-D}}} & \multicolumn{2}{c|}{\cellcolor{gray!10}\textit{\textbf{RGB-T}}} & \multicolumn{4}{c|}{\cellcolor{gray!10}\textit{\textbf{VDT}}} & \multicolumn{2}{c}{\cellcolor{gray!10}\textit{\textbf{Mamba-based}}} \\
			\cmidrule(lr){3-4} \cmidrule(lr){5-6} \cmidrule(lr){7-10} \cmidrule(lr){11-12}
			\multicolumn{2}{c|}{} & \makecell{DPA \\ \cite{dpa}} & \makecell{SwinNet \\ \cite{swinnet}} & \makecell{LSNet \\ \cite{lsnet}} & \makecell{CAFC \\ \cite{cafc}} & \makecell{TMNet \\ \cite{tmnet}} & \makecell{MFFNet \\ \cite{mffnet}} & \makecell{DWFPR \\ \cite{dwfpr}} & \makecell{PWRF \\ \cite{pwrf}} & \makecell{Samba \\ \cite{samba}} & \textbf{Ours} \\
			\midrule
			\multicolumn{2}{c|}{Params (M)} & 92.39 & 124.72 & 4.57 & - & - & 103.24 & - & - & 60.28 & 46.07 \\
			\midrule
			\multirow{3}{*}{\makecell{VDT-2048 \\ \cite{vdt2048}}} 
			& \(S_m \uparrow\) & 81.8 & 91.9 & 88.8 & 91.3 & 93.3 & 93.6 & \secondbest{93.8} & 93.2 & \secondbest{93.8} & \best{94.2} \\
			& \(F_m \uparrow\) & 45.9 & 72.9 & 74.3 & 85.4 & 89.7 & 90.1 & 90.1 & 91.0 & \secondbest{91.0} & \best{91.9} \\
			& \(E_m \uparrow\) & 68.6 & 89.6 & 92.0 & 97.7 & 98.9 & \secondbest{99.0} & \secondbest{99.0} & 98.8 & \secondbest{99.0} & \best{99.2} \\
			\bottomrule
		\end{tabular}
	}
\end{table}

\begin{figure}[t]
	\centering
	\includegraphics[width=0.84\linewidth]{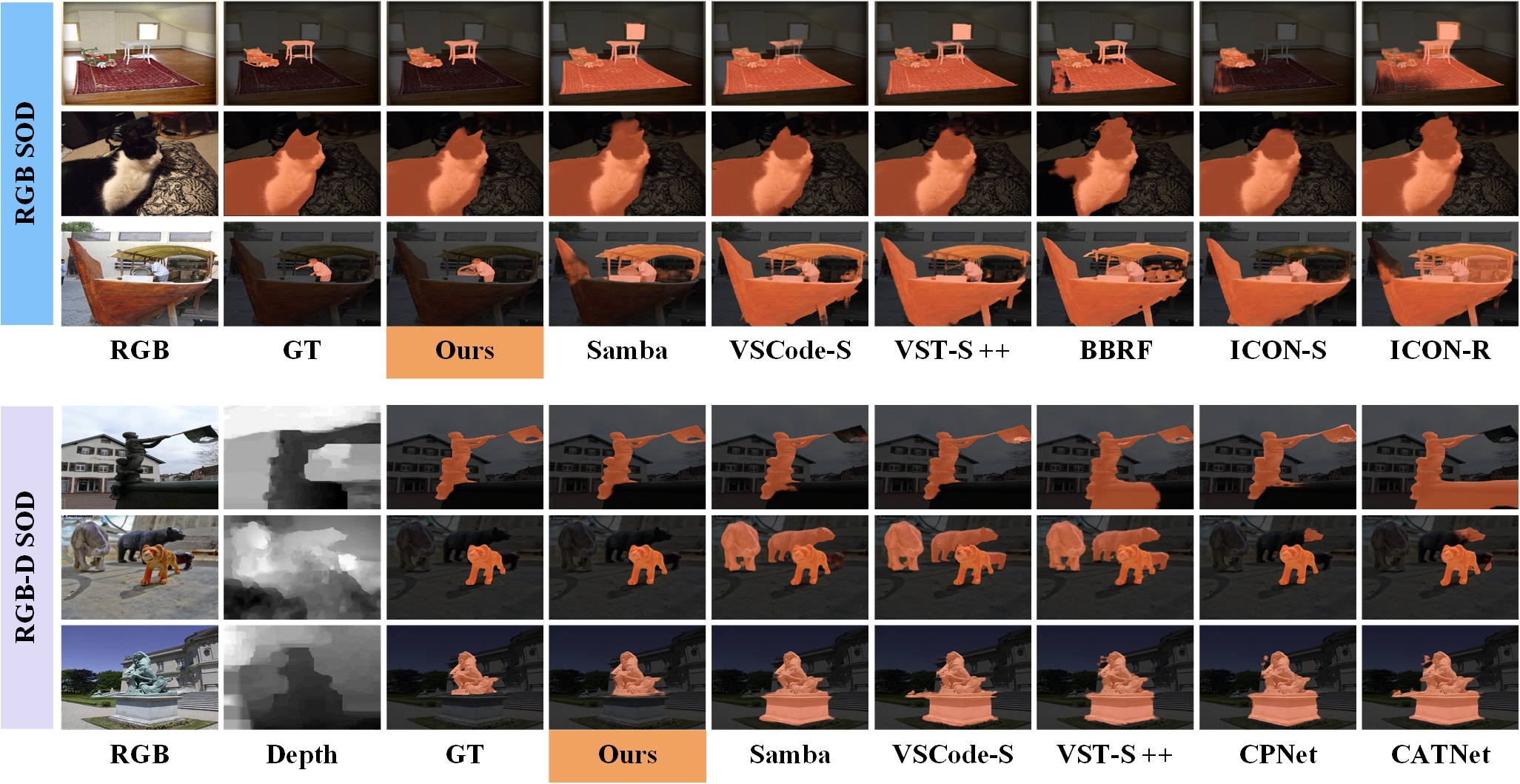}
	\caption{\textbf{Visual comparison against SOTA methods on RGB and RGB-D SOD tasks.} The top three rows show RGB scenarios, while the bottom three rows display RGB-D scenarios with corresponding depth maps.}
	\label{fig:visual_comparison}
\end{figure}

\subsection{Comparison with State-of-the-Art Methods}

\noindent\textbf{Quantitative Evaluation.} Since \modelname\ is designed as a heterogeneous hybrid framework for general SOD, we conduct extensive comparative experiments against existing state-of-the-art (SOTA) methods across five distinct tasks: RGB SOD, RGB-D SOD, RGB-T SOD, VSOD, and VDT SOD. The comprehensive quantitative results are presented in Tables~\ref{tab:main_results},~\ref{tab:rgbd_results},~\ref{tab:rgbt_results},~\ref{tab:vsod_results}, and~\ref{tab:vdt_results_comparison}. These comparisons demonstrate that \modelname\ consistently outperforms leading CNN-based, Transformer-based, and Mamba-based competitors across all benchmark datasets, validating the superior efficacy of our proposed architecture.

Specifically, for the \textbf{RGB SOD} task (Table~\ref{tab:main_results}), our model achieves the best performance on nearly all metrics while maintaining a significantly lower parameter count compared to heavy Transformer-based methods. For instance, on the challenging DUTS dataset, \modelname\ surpasses the heavy transformer ICON-S by 1.9\% in $S_m$ and 2.5\% in $F_m$, while using less than half of its parameters (43.23M vs. 94.30M). Even compared to the recent Mamba-based method Samba, our model achieves a consistent improvement of 0.4\%--1.6\% across five datasets, proving the advantage of our heterogeneous fusion over pure SSM modeling.

Regarding multi-modal tasks, including \textbf{RGB-D SOD} (Table~\ref{tab:rgbd_results}) and \textbf{RGB-T SOD} (Table~\ref{tab:rgbt_results}), \modelname\ sets new state-of-the-art standards. In RGB-D SOD, our method outperforms the strong competitor VST-S++ by a large margin (e.g., +2.2\% $S_m$ on NJUD) with only 31\% of its parameters (44.65M vs. 143.15M). Similarly, in RGB-T SOD, our method surpasses Samba on all three benchmarks, achieving a notable 0.7\% $S_m$ gain on the VT821 dataset. 

Furthermore, in video-based tasks (\textbf{VSOD} and \textbf{VDT}), our model demonstrates exceptional temporal and multi-modal modeling capabilities. As shown in Table~\ref{tab:vsod_results}, \modelname\ achieves significant gains over VSCode-S on the DAVSOD-easy dataset (+3.1\% $S_m$, +5.2\% $F_m$), highlighting its robustness in dynamic scenes. In the tri-modal \textbf{VDT SOD} task (Table~\ref{tab:vdt_results_comparison}), \modelname\ reaches an $S_m$ of 94.2\%, establishing the new state-of-the-art. Notably, compared to the strongest competitor \textit{Samba}, \modelname\ not only achieves superior segmentation accuracy across all five tasks but also benefits from a lower computational complexity. This indicates that our heterogeneous framework can effectively capture multi-modal and temporal saliency cues.

\noindent\textbf{Qualitative Evaluation.} To visually demonstrate the superiority of \modelname, we display qualitative comparisons in Fig.~\ref{fig:visual_comparison}. In RGB scenarios (top three rows), our model excels in handling complex topologies and low-contrast objects. Specifically, in the $1^{st}$ row, while competitors like Samba~\cite{samba} and VST-S++~\cite{vst++} fail to suppress the background gaps within the chair legs, \modelname\ cleanly delineates the object skeleton. In the $2^{nd}$ row, it accurately segments the black cat against a dark background, and in the $3^{rd}$ row, it recovers the complete boat structure despite severe internal occlusion, avoiding the fragmentation seen in ICON-R~\cite{icon}. In RGB-D scenarios (bottom three rows), \modelname\ effectively leverages depth cues to resolve visual ambiguities. For instance, in the $4^{th}$ row, it distinguishes the statue from the complex building texture where others fail. Similarly, in the $5^{th}$ and $6^{th}$ rows, our method consistently produces sharp predictions for the toy animals and fountain statue, robustness against similar foreground-background colors and challenging lighting, outperforming depth-aware competitors like CPNet~\cite{cpnet} and CATNet~\cite{catnet}. These results visually confirm that our heterogeneous hybrid paradigm effectively harmonizes global semantics with local details across diverse modalities. \textbf{More extensive visual comparisons across diverse scenarios can be found in the Supplementary Material.}

\subsection{Ablation Study}

To verify the contribution of each proposed component, we conduct thorough ablation studies across five representative datasets. As summarized in Table~\ref{tab:ablation_study}, our full architecture consistently outperforms all variants, validating the effectiveness of the heterogeneous hybrid design, the LFM, and the SGU operator.

\noindent\textbf{Effectiveness of Architecture (Settings A).} We evaluate the necessity of our heterogeneous hybrid design by comparing the full model with single-stream variants (A1, A2) and a naive dual-stream baseline (A3). The quantitative results demonstrate that relying solely on either the SSM-based stream (A1) or the CNN-based stream (A2) yields sub-optimal performance across all metrics. While the naive dual-stream baseline (A3) offers clear numerical improvements over both single streams by simply aggregating the two backbones, a noticeable performance gap persists compared to our full model. This confirms that the dual-stream architecture is inherently more effective than single streams, and our specific interaction design further maximizes their combined potential.

\noindent\textbf{Effectiveness of Feature Fusion Strategy (Settings B).} To validate our LFM, we replace it with standard operations: additive fusion (B1), simple concatenation (B2), and the widely-used cross-attention mechanism (B3). The metric comparisons consistently show that LFM achieves the highest scores across all five datasets. This numerical superiority indicates that our proposed dynamic fusion mechanism is a more effective strategy for aggregating heterogeneous features than conventional static fusion methods in dense prediction tasks.

\begin{table}[t]
	\centering
	\caption{Ablation study of our \modelname\ on five datasets. All metric values are in \%. \textbf{Bolded} results are the best.}
	\label{tab:ablation_study}
	\resizebox{\textwidth}{!}{
		\setlength{\tabcolsep}{2.5pt}
		\renewcommand{\arraystretch}{0.8}
		\begin{tabular}{l | ccc | ccc | ccc | ccc | ccc}
			\toprule
			\multirow{3}{*}{\textbf{Settings}} & \multicolumn{6}{c|}{\textit{\textbf{RGB SOD}}} & \multicolumn{6}{c|}{\textit{\textbf{RGB-D SOD}}} & \multicolumn{3}{c}{\textit{\textbf{VDT}}} \\
			\cmidrule(lr){2-7} \cmidrule(lr){8-13} \cmidrule(lr){14-16}
			& \multicolumn{3}{c|}{DUTS~\cite{duts}} & \multicolumn{3}{c|}{DUT-O~\cite{duto}} & \multicolumn{3}{c|}{NJUD~\cite{njud}} & \multicolumn{3}{c|}{NLPR~\cite{nlpr}} & \multicolumn{3}{c}{VDT-2048~\cite{vdt2048}} \\
			\cmidrule(lr){2-4} \cmidrule(lr){5-7} \cmidrule(lr){8-10} \cmidrule(lr){11-13} \cmidrule(lr){14-16}
			& \(S_m \uparrow\) & \(F_m \uparrow\) & \(E_m \uparrow\) & \(S_m \uparrow\) & \(F_m \uparrow\) & \(E_m \uparrow\) & \(S_m \uparrow\) & \(F_m \uparrow\) & \(E_m \uparrow\) & \(S_m \uparrow\) & \(F_m \uparrow\) & \(E_m \uparrow\) & \(S_m \uparrow\) & \(F_m \uparrow\) & \(E_m \uparrow\) \\
			\midrule
			\textbf{Ours (Full)} & \textbf{93.6} & \textbf{93.6} & \textbf{97.0} & \textbf{90.4} & \textbf{88.4} & \textbf{93.8} & \textbf{95.0} & \textbf{95.8} & \textbf{97.8} & \textbf{94.9} & \textbf{94.3} & \textbf{97.7} & \textbf{94.2} & \textbf{91.9} & \textbf{99.2} \\
			\midrule
			\rowcolor{orange!20} \multicolumn{16}{l}{\textit{\textbf{A: Ablation on Architecture}}} \\
			\midrule
			A1: VMamba Only & 92.0 & 91.7 & 96.4 & 89.5 & 87.1 & 93.6 & 94.3 & 94.7 & 97.3 & 94.2 & 93.5 & 97.4 & 91.4 & 87.4 & 97.9 \\
			A2: ConvNeXt Only & 84.8 & 82.5 & 88.3 & 83.3 & 78.3 & 86.0 & 82.2 & 81.7 & 86.7 & 84.7 & 79.9 & 89.1 & 86.8 & 80.9 & 93.7 \\
			A3: Dual Stream Baseline & 92.5 & 92.2 & 96.0 & 89.8 & 87.5 & 93.1 & 94.6 & 95.1 & 97.4 & 94.4 & 93.6 & 97.2 & 93.0 & 90.5 & 98.5 \\
			\midrule
			\rowcolor{cyan!10} \multicolumn{16}{l}{\textit{\textbf{B: Ablation on Feature Fusion Strategy (Replacing LFM)}}} \\
			\midrule
			B1: Additive Fusion & 92.7 & 92.6 & 96.1 & 89.9 & 87.8 & 93.2 & 94.7 & 95.3 & 97.5 & 94.5 & 93.8 & 97.3 & 93.5 & 90.9 & 98.7 \\
			B2: Concatenation & 93.0 & 92.8 & 96.3 & 90.1 & 88.0 & 93.4 & 94.8 & 95.5 & 97.6 & 94.7 & 94.0 & 97.5 & 93.7 & 91.2 & 98.9 \\
			B3: Cross-Attention & 93.2 & 93.1 & 96.7 & 90.2 & 88.1 & 93.5 & 94.7 & 95.3 & 97.5 & 94.7 & 94.0 & 97.4 & 94.0 & 91.6 & 99.0 \\
			\midrule
			\rowcolor{green!10} \multicolumn{16}{l}{\textit{\textbf{C: Ablation on Upsampling Strategy (Replacing SGU)}}} \\
			\midrule
			C1: Bilinear & 92.9 & 92.8 & 96.4 & 89.8 & 87.7 & 93.1 & 94.3 & 95.0 & 97.0 & 94.2 & 93.5 & 96.9 & 93.2 & 90.9 & 98.4 \\
			C2: Transposed Conv & 93.2 & 93.1 & 96.7 & 90.1 & 88.0 & 93.4 & 94.6 & 95.4 & 97.4 & 94.6 & 93.9 & 97.3 & 93.5 & 91.3 & 98.7 \\
			\bottomrule
		\end{tabular}
	}
\end{table}

\noindent\textbf{Effectiveness of Upsampling Strategy (Settings C).} The Saliency-Guided Upsampling (SGU) operator is designed to better propagate semantics across multi-level features during resolution restoration. Comparing SGU with standard bilinear interpolation (C1) and transposed convolution (C2), we observe consistent quantitative gains across all multi-modal tasks. These continuous metric improvements confirm that our specialized SGU effectively maintains semantic consistency during the upsampling process, leading to more accurate overall detection performance.

\section{Conclusion}
In this paper, we propose \modelname\, a heterogeneous hybrid framework integrating VMamba and ConvNeXt for general SOD, designed to bridge the inherent spectral biases of single-paradigm networks. By revealing the complementary frequency preferences of VMamba and ConvNeXt, we developed a state-stimulus paradigm within the liquid fusion for dynamic feature aggregation. Complemented by the saliency-guided upsampling operator for effective feature propagation and boundary recovery, \modelname\ consistently establishes new state-of-the-art benchmarks across five general SOD tasks. Future work will focus on extending this liquid fusion paradigm to broader dense prediction scenarios and optimizing the architecture for lightweight edge applications.

\section*{Acknowledgements}
The paper was supported in part by the National Natural Science Foundation of China under Grant 62571068, 62306048; in part by Science and Technology Development Fund, Macao SAR (Grant No:
0054/2025/RIB2); in part by the Qing Lan Project of Jiangsu Universities; in part by the Major Program of Jiangsu Higher Education Institutions Basic Science (Natural Science) Research under Grant 25KJA520001; in part by Changzhou Applied Basic Research Fund Project under Grant CJ20242060, CQ20230092, CJ20235036.

%
%
\bibliographystyle{splncs04}
\bibliography{main}
\end{document}